\pdfoutput=1
\documentclass[11pt]{article}

\usepackage{ACL2024}

\usepackage{hyperref}

\usepackage{times}
\usepackage{latexsym}
\usepackage{arydshln}
\usepackage{graphicx}
\usepackage{subcaption}
\usepackage{booktabs,arydshln}
\usepackage{amsmath}
\usepackage{enumitem}
\usepackage{multirow}
\usepackage{cleveref}
\usepackage[export]{adjustbox}
\usepackage[utf8]{inputenc}
\usepackage{xurl}
\usepackage{tabularray}
\usepackage{siunitx}
\usepackage{amssymb}% http://ctan.org/pkg/amssymb
\usepackage{pifont}% http://ctan.org/pkg/pifont
\newcommand{\cmark}{\ding{51}}%
\newcommand{\xmark}{\ding{55}}%

\newcommand{\circa}{{\raise.17ex\hbox{$\scriptstyle\sim$}}}

\usepackage{array}
\newcolumntype{L}[1]{>{\raggedright\let\newline\\\arraybackslash\hspace{0pt}}m{#1}}
\newcolumntype{C}[1]{>{\centering\let\newline\\\arraybackslash\hspace{0pt}}m{#1}}
\newcolumntype{R}[1]{>{\raggedleft\let\newline\\\arraybackslash\hspace{0pt}}m{#1}}

\newcommand\nnfootnote[1]{%
  \begin{NoHyper}
  \renewcommand\thefootnote{}\footnote{#1}%
  \addtocounter{footnote}{-1}%
  \end{NoHyper}
}

\usepackage[T1]{fontenc}
\usepackage[utf8]{inputenc}

\usepackage{microtype}

\usepackage[colorinlistoftodos]{todonotes}
\usepackage{cleveref}
\usepackage{listings}

\makeatletter
\newcommand*\iftodonotes{\if@todonotes@disabled\expandafter\@secondoftwo\else\expandafter\@firstoftwo\fi} 
\makeatother

\crefname{section}{\S}{\S\S} % {singular}{plural}
\Crefname{section}{\S}{\S\S} % \Cref{...} for capitalized version
\crefname{table}{Tab.}{Tables}
\crefname{figure}{Fig.}{Figures}
\crefname{algorithm}{Algorithm}{}
\crefname{equation}{eq.}{}
\crefname{appendix}{App.}{}
\crefname{lstlisting}{listing}{listings}
\Crefname{lstlisting}{Listing}{Listings}
\crefformat{section}{\S#2#1#3} 

\definecolor{KUPetrol}{RGB}{0,120,148} % green/blue

\definecolor{KUBlue}{RGB}{33,92,175} % blue
\definecolor{KUGreen}{RGB}{98,115,19} % green
\definecolor{KUPurpleDark}{RGB}{140,10,89} % purple
\definecolor{KUPurple}{RGB}{163,7,116} % purple
\definecolor{KUGray}{RGB}{111,111,111} % gray
\definecolor{KURed}{RGB}{183,53,45} % red
\definecolor{KUPetrol}{RGB}{0,120,148} % green/blue
\definecolor{KUBronze}{RGB}{142,103,19} % bronze

\newtoggle{color-macro}
\settoggle{color-macro}{true} % set to false to disable color macros

\iftoggle{color-macro}{\colorlet{MacroColor}{KUPetrol}
}{
\colorlet{MacroColor}{black}
}

\iftoggle{color-macro}{
\colorlet{TokenColor}{KUBronze}
}{
\colorlet{TokenColor}{black}
}

\iftoggle{color-macro}{
\colorlet{MathSubColor}{KUPurple}
}{
\colorlet{MathSubColor}{black}
}

\iftoggle{color-macro}{
\colorlet{RedditColor}{black}
}{
\colorlet{RedditColor}{black}
}

\iftoggle{color-macro}{
\colorlet{SchwartzProbColor}{KUBlue}
}{
\colorlet{SchwartzProbColor}{black}
}

\iftoggle{color-macro}{
\colorlet{IColor}{KUGray}
}{
\colorlet{IColor}{black}
}

\usepackage[T1]{fontenc}

\usepackage{booktabs} % for "\addlinespace" macro

\usepackage{amsmath,amsfonts,bm}

\def\eqref#1{equation~\ref{#1}}
\def\1{\bm{1}}

\DeclareMathAlphabet{\mathsfit}{\encodingdefault}{\sfdefault}{m}{sl}
\SetMathAlphabet{\mathsfit}{bold}{\encodingdefault}{\sfdefault}{bx}{n}

\definecolor{codegreen}{rgb}{0,0.6,0}
\definecolor{codegray}{rgb}{0.5,0.5,0.5}
\definecolor{codepurple}{rgb}{0.58,0,0.82}
\definecolor{backcolour}{rgb}{0.97,0.97,0.95}

\lstdefinestyle{mystyle}{
    backgroundcolor=\color{backcolour},   
    commentstyle=\color{codegreen},
    keywordstyle=\color{magenta},
    numberstyle=\tiny\color{codegray},
    stringstyle=\color{codepurple},
    basicstyle=\ttfamily\footnotesize,
    breakatwhitespace=true,         
    breaklines=true,                 
    captionpos=b,                    
    keepspaces=true,                 
    numbers=left,                    
    numbersep=5pt,                  
    showspaces=false,                
    showstringspaces=false,
    showtabs=false,                  
    tabsize=2
}

\title{Can Community Notes Replace Professional Fact-Checkers?}

\author{
Nadav Borenstein$^{\ast}$ \quad
Greta Warren$^{\ast}$ \quad
Desmond Elliott \quad
Isabelle Augenstein \\
University of Copenhagen \quad \\
{\tt \href{mailto:nb@di.ku.dk}{nb@di.ku.dk}} \quad
{\tt \href{mailto:grwa@di.ku.dk}{grwa@di.ku.dk}} \quad \\
{\tt \href{mailto:de@di.ku.dk}{de@di.ku.dk}}  \quad 
{\tt \href{mailto:augenstein@di.ku.dk}{augenstein@di.ku.dk}} \quad 
}

\begin{document}
\maketitle
% \listoftodos

% \setlength{\parskip}{0cm plus0mm minus0mm}

\begin{abstract}

Two commonly employed strategies to combat the rise of misinformation on social media are (i) fact-checking by professional organisations and (ii) community moderation by platform users. Policy changes by Twitter/X and, more recently, Meta, signal a shift away from partnerships with fact-checking organisations and towards an increased reliance on crowdsourced community notes. However, the extent and nature of dependencies between fact-checking and \emph{helpful} community notes remain unclear. To address these questions, we use language models to annotate a large corpus of Twitter/X community notes with attributes such as topic, cited sources, and whether they refute claims tied to broader misinformation narratives. Our analysis reveals that community notes cite fact-checking sources up to five times more than previously reported. Fact-checking is especially crucial for notes on posts linked to broader narratives, which are \textit{twice} as likely to reference fact-checking sources compared to other sources. Our results show that successful community moderation relies on professional fact-checking and highlight how citizen and professional fact-checking are deeply intertwined.

\nnfootnote{* Equal contribution.}

% We discuss the implications of these findings for the future of fact-checking.

% We discover that effective community moderation depends on professional fact-checking

% The increasing prevalence of misinformation on social media platforms has motivated various approaches to debunking false information.  Policy changes by Twitter/X and more recently, Meta, have signalled a shift away from partnerships with fact-checking organisations and increased reliance on crowdsourced community notes. However, it is unclear (i) whether community notes represent an adequate stand-alone solution and (ii) whether notes targeting certain types of misinformation are more reliant on fact-checkers than others. 
% To answer these questions, we analyse and annotate a large corpus of Twitter/X community notes.
% Our findings indicate that successful community moderation is reliant on professional fact-checking: community notes that link to fact-checking articles are deemed more helpful than those that do not. \textcolor{red}{Say that they rely on these sources at least 4 times more than was previously reported.}
% We also find that professional fact-checking is particularly vital for community notes attached to posts belonging to broader misinformation narratives (e.g., conspiracy theories).
% We discuss the implications of these findings for the future of fact-checking.

\end{abstract}
\everypar{\looseness=-1}

\section{Introduction}
\label{sec:introduction}

\begin{figure}[t]
    \centering
    \includegraphics[width=0.95\columnwidth]{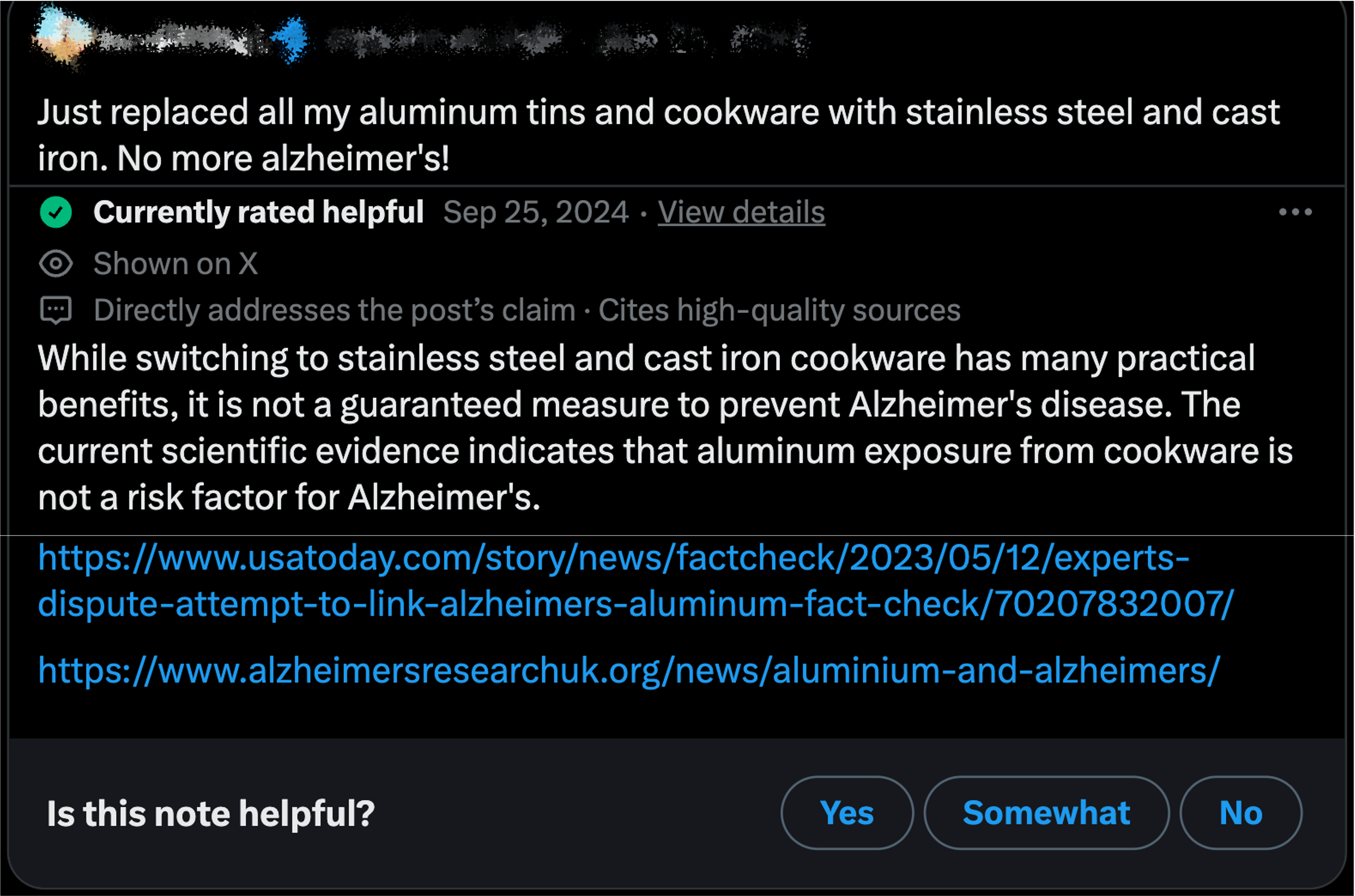}
    \caption{An example of a community note. Notice the fact-checking link and rating.}
    \label{fig:notes_sample}
\end{figure}

%% put in implications section
% In addition to verifying claims, in recent years many fact-checking organisations have also assumed a wider role in combatting misinformation spread, conducting long-term investigative journalism projects and citizen media literacy programs \citep{juneja2022human}. 
The proliferation of misinformation on social media \citep{arnold2020onlineFC,diakopoulos2020computational}, along with the rise of generative AI \citep{augenstein2024factualityLLMs} have led to increasing concerns about people's ability to access trustworthy and credible information, leading to potential harms to public health \citep{ijerph19095321}, democracy, and political stability 
% \citep{McKay2021misinfodemocracy,reglitz2022fake}.
\citep{reglitz2022fake}. 
Fact-checkers play a crucial role in combatting misinformation \citep{graves2017anatomy}, and in recent years, have partnered with social media platforms, e.g., Meta, YouTube, and TikTok, to tackle its spread on these platforms. However, due to the scale of misleading content shared online, community moderation (e.g., options to flag potential misinformation, group/server moderators) is often employed in parallel \citep{morrow2022emerging}, as a complementary approach (e.g., \citep{YouTubeMisinfo, tiktoknotes}; see also the practice of \textit{snoping} \citep{pilarski_community_2024}).  
The expansion of fact-checking projects in the last decade \citep{lauer2024growfactcheck}, alongside their broader initiatives to curb misinformation (e.g., citizen media literacy programmes \citep{juneja2022human}) have been aided by partnerships with social media platforms such as Meta and Google \citep{graves2020infrastructure}, which fund independent fact-checking agencies to fact-check potentially false claims on their platform.\footnote{Fact-checkers provide a judgment of claim veracity and exert no influence on the platforms' content moderation policies \citep{PoynterMeta2025}.} 
However, political pressure and accusations of bias and censorship, %from politicians and social media platforms, 
and most recently, Meta's announcement of its plans to end its partnerships with fact-checkers in the U.S. and implement a community moderation model \citep{MetaFactChecking2025}, threatens the financial stability of fact-checking organisations, and hence, their ability to keep up with the increasing volume and sophistication of misinformation spread \citep{Duke2024FactCheckingSputters,IFCN2024report}.

% Community moderation has been proposed as a means of scaling up fact-checking \citep{martel2024crowdmisinfo}, and partly addressing the challenges of establishing cross-partisan trust in fact-checking \citep{poynter2019republicans}. One of the most notable examples of such a system is Twitter/X, which commenced its pilot Community Notes programme (then called Birdwatch) in January 2021, and later launched it to the public in October 2022 \citep{TwitterBirdwatch2021}. 

Meta's recent policy shift also implies that these two strategies (fact-checking and community notes) are independent and in opposition, rather than two complementary strategies of tackling online misinformation.
In this paper, \textit{we examine Twitter/X community notes as a case study to understand how fact-checking is used in community notes}. Specifically, we investigate the following two questions:  \textbf{(RQ1) To what extent do community notes rely on the work of professional fact-checkers?} and \textbf{(RQ2) What are the traits of posts and notes that rely on fact-checking sources?} 
Studying the relationship between fact-checking and community notes is vital for understanding the shared role of expert and citizen-driven fact-checking in the global information ecosystem.

% Answering these questions can shed light on the long-term implications of this move. 

We find that at least 1 in 20 community notes rely explicitly on the work of professional fact-checkers, while this reliance is higher still for high-stakes topics such as health and politics. Our experiments also show that fact-checking is vital for debunking misleading content linked to broader narratives or conspiracy theories. These findings imply that high-quality community notes cannot be produced independently of professional fact-checking. They further suggest that the pressure on fact-checkers exerted by platforms and politicians by defunding and discrediting fact-checking organisations will have corrosive effects on the quality of notes and destructive implications for information integrity more widely.

\section{Background}
\label{sec:related_work}
Due to space constraints, only the most relevant related work is provided here. Additional relevant studies and background can be found in \Cref{app:extended_background}.

\subsection{Community notes}

Community moderation has been proposed as a means of addressing the scalability \citep{martel2024crowdmisinfo} and cross-partisanship trust \citep{poynter2019republicans} challenges associated with fact-checking. 
Twitter/X's Community Notes programme (piloted in 2021 and publicly launched in October 2022 \citep{TwitterBirdwatch2021}) is a notable example of such a system.
Any platform user may volunteer as a Community Notes contributor, although they must achieve a particular `rating impact score' before they can write notes \citep{TwitterRatingWritingImpact}.
Notes that achieve a `helpful' rating appear underneath the post, explaining why the post is misleading (see \cref{fig:notes_sample}). To be rated `helpful', a note must receive similar levels of helpfulness rating from users with diverse viewpoints \citep{TwitterNoteRanking}.

\subsubsection{Characteristics of Community Notes}

A small but growing body of work has analysed Twitter/X's Community Notes dataset, focusing on the targets, sources, and limitations of notes. 

\noindent \textbf{Targets of notes.}
Community notes tend to focus on misleading posts from large accounts %more exclusively than snopes do 
\citep{pilarski_community_2024}, focusing on posts that lack important content or present unverified claims as facts \citep{prollochs2022community,drolsbach_diffusion_2023}. 
% More recent analyses, however, have also shown that the more influential the account, the lower the level of consensus among users \citep{prollochs2022community}. 

\noindent \textbf{Sources in notes.}
Analyses have showed that notes were rated more helpful if they link to `trustworthy' sources 
%\citep{maldita2025} 
and that the majority of sources cited by notes were `trustworthy'
% \footnote{rated to be of high factuality by \href{https://mediabiasfactcheck.com/}{Media Bias Fact Check}.}
left-leaning news outlets \citep{prollochs2022community}. %pointing to a potential bias in the platform's community fact-checking 
A recent study finds that 55\% of URLs used in notes were related to news websites, 18\% to research, 9\% to social media, 9\% to encyclopedic sources, but \textit{just 1\%} to fact-checking sources \citep{kangur_who_2024}.

\noindent \textbf{Limitations of notes.}
Only 11\% of submitted notes reach `helpful' status (i.e., shown to users) by achieving a cross-perspective \citep{renault_collaboratively_2024, wirtschafter2023future}, and the time frame for notes to reach the algorithm's required agreement level (15.5 hours on average) limits its capacity to halt misinformation spread \citep{renault_collaboratively_2024}. Posts related to partisan issues are particularly affected by these challenges \citep{allen2022partisan}. Additional concerns about the notes' efficacy highlight their indifference to the expertise needed for certain claims and reliance on subjective helpfulness rather than objective facts, free labour and inadequate support and guardrails regarding explicit content \citep{Gilbert_2025}. 

Our work provides novel insights into the targets, sources and limitations of community notes by shedding light on the relationship between notes and professional fact-checking.
Closest to our work is a recent analysis of community notes written in 2024 by the fact-checking organisation \citet{maldita2025}, who also studied the reliance of community notes on professional fact-checkers. They discover that fact-checking organisations are widely used as a reference by notes' authors. The current work provides a more fine-grained analysis by studying the extent to which fact-checking sources form the basis of note-writers' efforts to counter misinformation and identifying the strategies they employ.

\section{Dataset}
\label{sec:dataset}
\begin{figure*}[ht]
    \centering
    \includegraphics[width=\textwidth]{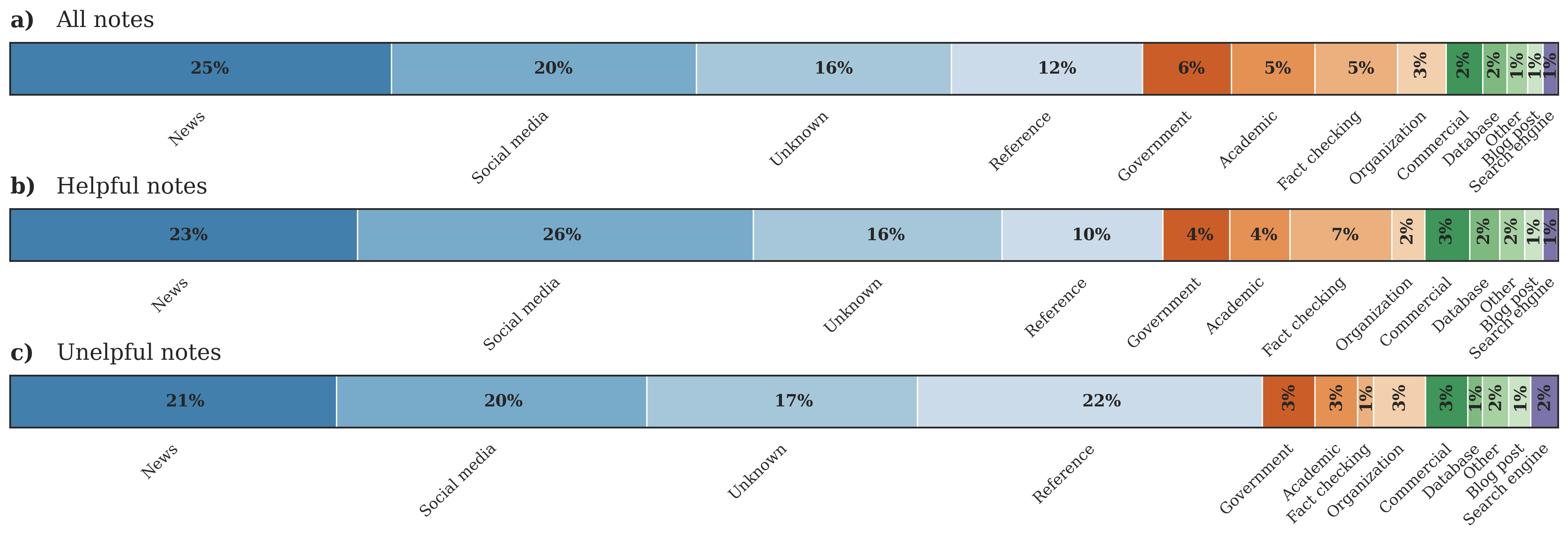}
    \caption{The categories of links used by Community notes' authors as a source. a) all community notes; b) Community notes rated as `helpful'; c) community notes rated as `unhelpful'. Notice the `fact-checking' category.}
    \label{fig:link_types}
\end{figure*}

We download files containing all community notes and their metadata from the official website,\footnote{\url{https://communitynotes.x.com/guide/en/under-the-hood/download-data}} which amounts to 1.5M notes authored between January 28th 2021 and January 6th 2025. Of these, a total of 135K are rated by the community as `Helpful', 51K are rated `Not helpful', and 1.3M are unpublished, i.e., did not receive enough community ratings to reach a verdict. See \cref{fig:notes_per_month} in \cref{app:additional_material} for statistics.

We filter the notes as follows. First, we remove 526K non-English notes, which we identify by applying the language detection library fast-langdetect.\footnote{\url{https://github.com/LlmKira/fast-langdetect}} Then, we further filter 268K `unnecessary' notes---notes attached to tweets that are classified by the community as `not misleading'. Finally, to focus only on notes that are used to address misinformation, we filter out 44K notes that contain one of the words `ad', `spam', or `phishing'. Following these filtration steps, we are left with a dataset containing 664K notes.

The next step involves categorising the sources that the note authors use to support their claims. First, we use regex to extract all the URLs found in the notes. Importantly, a single note can include multiple external URLs as evidence. See \cref{tab:top_domains} in \cref{app:additional_material} for a list of the top-100 most common domains. We classify each URL in our dataset of 664K notes into one of 13 categories (detailed in \cref{fig:link_types}) using the pipeline described below.

\begin{enumerate}[topsep=0pt,itemsep=-1ex,partopsep=1ex,parsep=1ex] % I don't know if this is allowed
    \item Check whether the domain name of the URL is found in a manually curated list of domains of professional fact-checking organisations (See \cref{tab:fact_check_orgs} in \cref{app:additional_material} for the full list). If so, classify the URL as `fact-checking'.
    \item Otherwise, search for paraphrases of the word `fact-check' in the URL,\footnote{These URLs mostly link to the fact-checking divisions of news outlets, e.g., \url{https://apnews.com/article/fact-checking-909101991741}.} and classify it as `fact-checking' if a match was found.
    \item Otherwise, check whether the domain name is found in \cref{tab:top_domains}, which the authors of this paper manually annotated.
    \item Otherwise, use GPT-4\footnote{Version \texttt{gpt-4o-2024-08-06}.} to classify the domain name into one of the 13 categories. \Cref{lst:prompt_link} in \cref{app:reproducibility} details the prompt we used. 
    \item Finally, if  GPT-4 fails or outputs an unknown category, label the URL as `unknown'.
\end{enumerate}

\noindent Using this pipeline, we successfully classify 95\% of the URLs to one of the 13 categories.

Moreover, we further subsample the notes for performing the in-depth analysis required for answering RQ2 (\cref{sec:analysis_rq2}). From the notes rated as `helpful', we sample 3.5K notes with a `fact-checking' source and a random sample of 22K additional notes. We then used web crawling to scrape the text of the posts to which these notes were attached. We name this subset $\mathcal{S}_\text{text}$ for simplicity.

\section{Analysis}
\label{sec:analysis}

\begin{figure}
    \centering
    \includegraphics[width=\columnwidth]{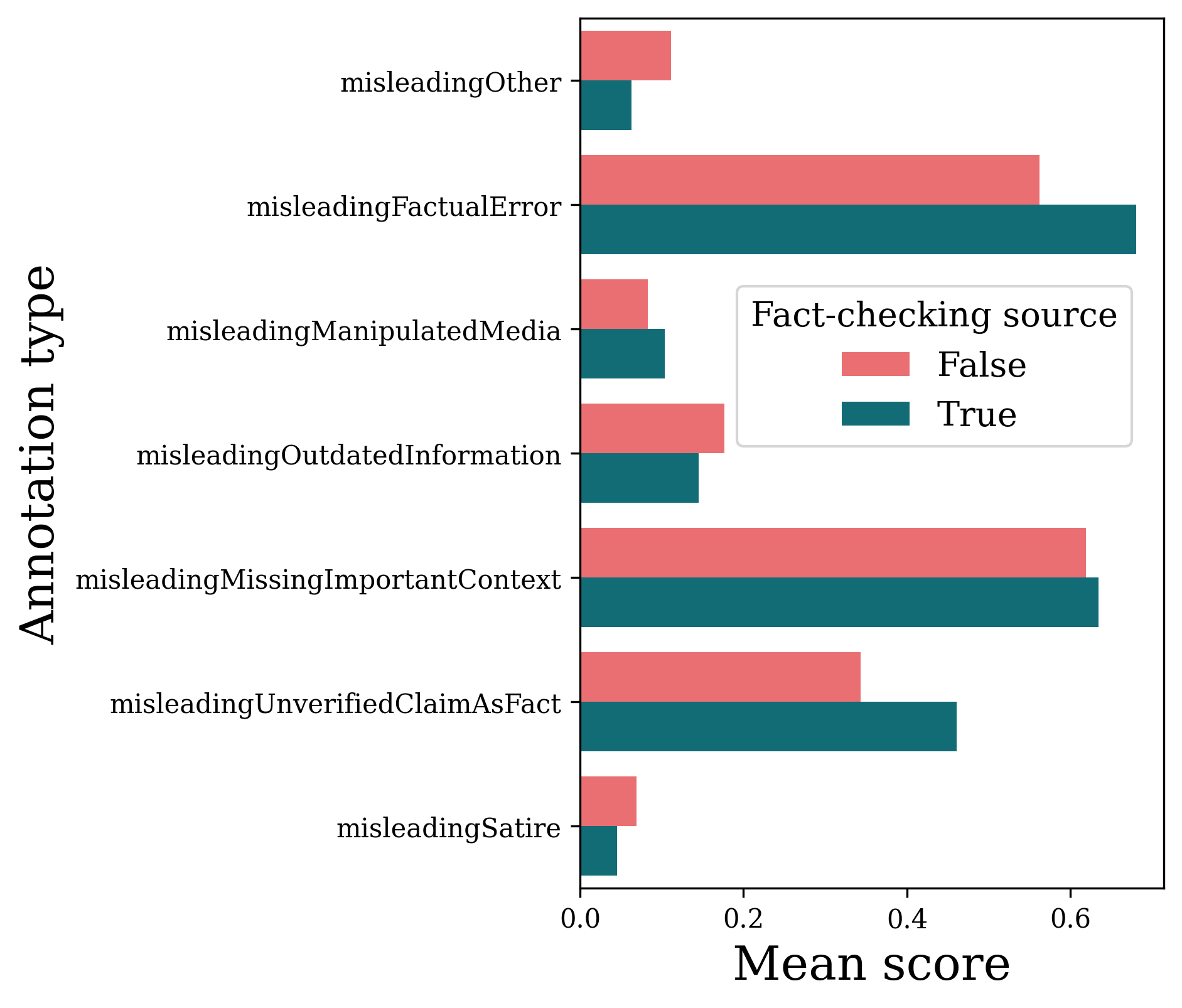}
    \caption{Mean scores of community annotations of misleading posts.}
    \label{fig:annotations}
\end{figure}

We analyse the dataset prepared in \cref{sec:dataset} to answer the two research questions defined in \cref{sec:introduction}.

\subsection{RQ1: To what degree do community notes rely on fact-checkers?}
\label{sec:analysis_rq1}

According to \Cref{fig:link_types}.a at least 5\% of all English community notes contain an external link to professional fact-checkers. This number grows to 7\% when only considering notes rated as `helpful' (\Cref{fig:link_types}.b). Conversely, only 1\% of notes rated as `not helpful' contain a fact-checking source (\Cref{fig:link_types}.c). These figures are significantly larger than what was reported in some previous studies (1.2\% \citep{kangur_who_2024}), possibly because \citet{kangur_who_2024} utilise a smaller dataset of fact-checking agencies and classify fact-checking divisions of popular journals as `news'. The results imply that notes incorporating fact-checking sources are generally considered more helpful. 

We further assess whether notes with fact-checking sources are perceived to be of higher quality by analysing individual user ratings of notes both with and without such sources. Specifically, we collect user ratings for a balanced
(i.e., including of a fact-checking source or not) sample of 20K notes rated by at least 50 users, 
% , with half containing a link to professional fact-checking and the other half without.
and calculated the average ratings for the notes. As can be seen in \cref{fig:notes_individual_ratings} in \cref{app:additional_material}, community notes with fact-checking sources are generally rated higher than their counterparts. Interestingly, while notes with fact-checking links are more likely to be regarded as having a good source (higher \textit{HelpfulGoodSources}), they are also more likely to be rated as \textit{notHelpfulSourcesMissingOrUnreliable}.  \cref{tab:notes_with_bad_source.} in \cref{app:additional_material} contains a sample of such notes.

\subsection{RQ2: What are the traits of posts and notes that rely on fact-checking sources?}
\label{sec:analysis_rq2}

\begin{figure}[!t]
    \centering
    \includegraphics[width=1\columnwidth]{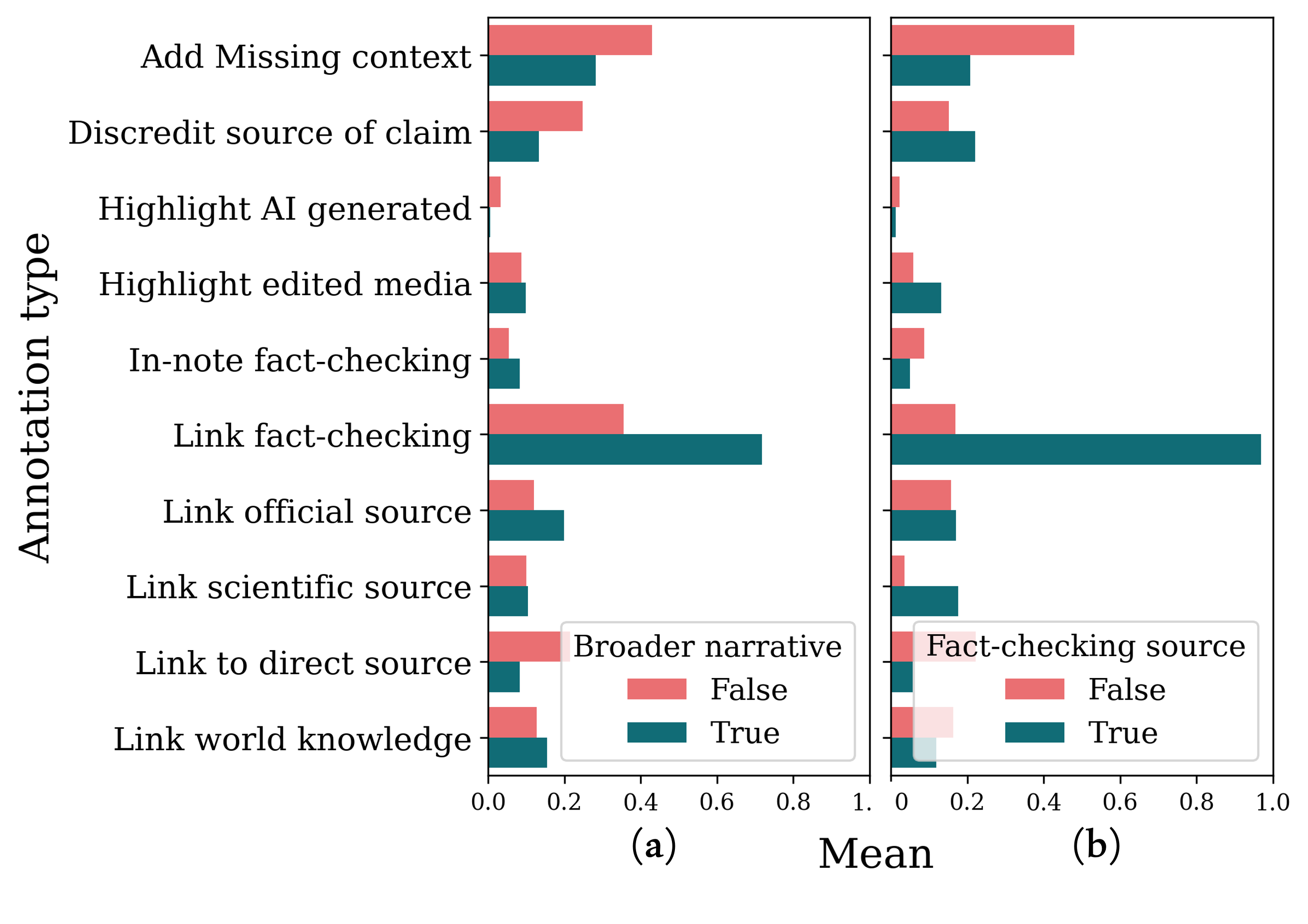}
    \caption{(a) strategies in debunking claims related to broader narratives. (b) the different ways in which fact-checking sources are used to debunk claims.}
    \label{fig:manual_annotation}
\end{figure}
% \begin{table}[h]
%     \centering
%     \begin{tabular}{p{2cm}lcc}
     
%        & & \multicolumn{2}{c}{Fact-check source} \\
        
%       & & Yes & No \\
%       \cline{3-4}
%         \multirow{2}{*}{\shortstack[l]{Related to a\\conspiracy} } & Yes & 0.216 & 0.112 \\
%         & No & 0.279 & 0.39 \\
%     \end{tabular}
%     \caption{Your table caption here}
%     \label{tab:fact_check}
% \end{table}

\def\arrvline{\hfil\kern\arraycolsep\vline\kern-\arraycolsep\hfilneg}

\begin{table}[!t]
% \fontsize{9}{9}\selectfont
    \centering
    \begin{tabular}{llc|c}
     
       & & \multicolumn{2}{c}{FC source} \\ 
       
      & & \cmark & \xmark \\
       \cmidrule(l){3-4}
      % \cmidrule(r){3-3}\cmidrule(l){4-4}
       \multirow{2}{*}{\rotatebox[origin=r]{90}{\parbox[r]{0.5cm}{\centering Conspi-racy}}} & \cmark \arrvline &  22\% & 11\% \\
       \cmidrule(l){2-4}
        & \xmark \arrvline & 28\% & 39\% \\
    \end{tabular}
    \caption{Percentage of samples related to a broader narrative or conspiracy vs. have a fact-checking source.}
    \label{tab:conspiracies_model_results}
\end{table}

\begin{figure}[!t]
    \centering
    \includegraphics[width=1\columnwidth]{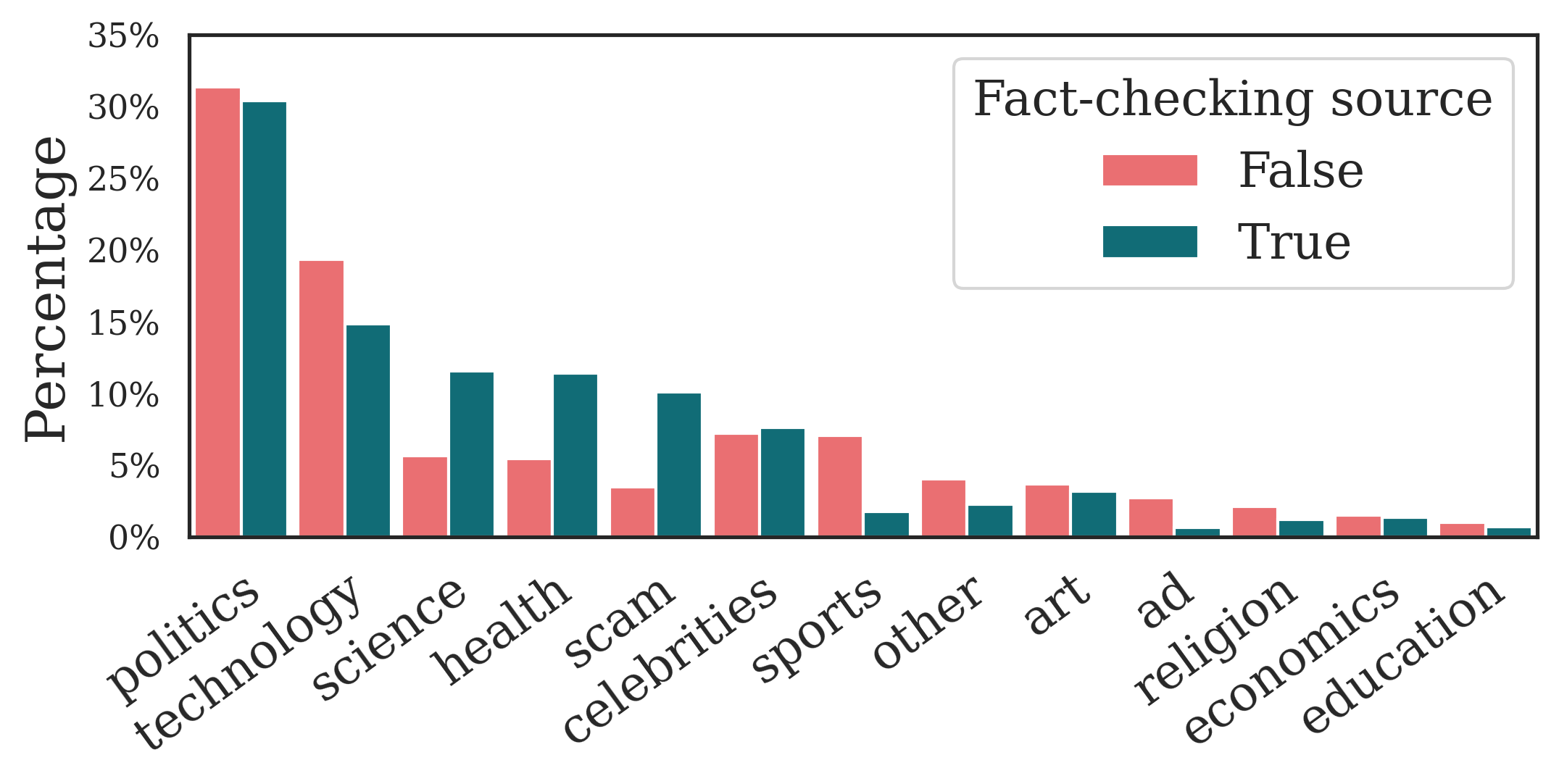}
    \caption{Distribution of notes' topics, with and without a fact-checking source.}
    \label{fig:topics}
\end{figure}

We begin by performing a topic analysis, comparing topics of posts whose notes reference fact-checking sources to those citing other sources. To this end, we apply a strong zero-shot text classification model\footnote{\url{https://huggingface.co/r-f/ModernBERT-large-zeroshot-v1} with default settings.} to our $\mathcal{S}_\text{text}$ subset by classifying spans of the form ``\texttt{Tweet:<POST TEXT>; Note <NOTE TEXT>}'' into one of 13 classes. The authors manually evaluated the quality of the classification results and considered it satisfactory, with the model predicting the correct class in $90\%$ of the cases. Most of the incorrect predictions involved the `technology' category, with sentences such as ``\textit{this is a fake image that was created with AI}'' incorrectly labelled as `technology'. Notably (\cref{fig:topics}), fact-checking sources are more likely to be included in posts related to high-stakes issues such as health, science, and scams and less likely to be included in posts on tech or sports.

We then analyse annotations (binary attributes explaining the warrant for the note) by community note authors.
% When writing a note, the author labels the original post with 
\cref{fig:annotations} contains the full breakdown of annotations for notes with and without fact-checking sources. Notes containing a link to fact-checking sources are overrepresented in posts where unverified information is presented as a fact or when the post contains a factual error. Conversely, they are under-represented in posts with outdated information or satirical content. \cref{tab:community_annotation_example} in \cref{app:additional_material} contains a sample of such notes. 

These results indicate that community note-writers adapt their strategies based on the stakes and scope of the claim, and the depth of research needed to counter misinformation. We hypothesise that they are more likely to rely on external fact-checking when refuting complex or unverifiable claims \citep{wuehrl-etal-2024-makes}, as well as claims related to conspiracy theories\footnote{For example, the claim ``Michelle Obama is a male''.} or broader narratives\footnote{These are tweets that perpetuate broader misinformation narratives but do not necessarily contain an explicit conspiracy theory, e.g., the false claim ``most immigrants remain firmly dependent on Welfare''.} which cannot be fully addressed in the scope of a note. Conversely, claims involving misleading media can often be debunked with examples alone, making fact-checking sources unnecessary. To investigate this hypothesis, the authors of this paper manually annotated 400 \texttt{<}post, note\texttt{>} pairs from $\mathcal{S}_\text{text}$ with attributes related to the complexity of the claims and how community notes address them. (see \cref{app:manual_annotation_setup} for annotation guidelines). The results (\cref{fig:manual_annotation}.a) support our hypothesis. Claims related to broader narratives or conspiracy theories\footnote{We condense broader narratives and conspiracy theories together for simplicity. Similar trends can be seen when they are analysed separately.} are much more likely to include a link to a fact-checking source.
In contrast, other types of claims are more likely to be addressed by providing missing context or by invalidating the credibility of the claim's source. 
Additionally, \cref{fig:manual_annotation}.b depicts the different ways in which fact-checking sources are used to debunk claims. It demonstrates how such sources are rarely used to provide missing context but rather focus on discrediting sources of claims and providing scientific evidence.

We extend the manual annotation to an LLM-based analysis of 8K balanced \texttt{<}post, note\texttt{>} pairs from $\mathcal{S}_\text{text}$. We task OpenAI's GPT-4\footnote{Version \texttt{gpt-4o-2024-08-06}.} with determining whether a pair relates to a broader narrative or a conspiracy theory. \Cref{lst:prompt_conspiracy} in \cref{app:reproducibility} details the prompt we used. To evaluate model accuracy, two authors independently labelled 100 balanced pairs, achieving an agreement rate of $0.88$ and resolving disagreements through discussion. The model attained an 
$F_1$ score of $0.85$---strong performance for this challenging task. The results (\cref{tab:conspiracies_model_results}) support our hypothesis: \texttt{<}post, note\texttt{>} pairs related to a broader narrative or conspiracy theory are \textit{twice} as likely to cite fact-checking sources compared to other sources. In contrast, other pairs are nearly 30\% less likely to do so. These findings also highlight the prevalence of such claims and further underscore the importance of fact-checking in combating complex misinformation narratives.

\section{Conclusion}
\label{sec:conclusion}
In this work, we annotate a large corpus of Twitter community notes with attributes such as topic, cited sources, and whether they refute claims tied to broader misinformation narratives. We find that effective community moderation depends on professional fact-checking to an extent far greater than previously reported. We find that community notes linked to broader narratives or conspiracy theories are particularly reliant on fact-checking.

Our results reveal that community notes and professional fact-checking are deeply interconnected—fact-checkers conduct in-depth research beyond the reach of amateur platform users, while community notes publicise their work. The move by platforms to end their partnerships and funding for fact-checking organisations will hinder their ability to fact-check and pursue investigative journalism, which community note writers rely on. This, in turn, will limit the efficacy of community notes, especially for high-stakes claims tied to broader narratives or conspiracies.

% This indicates that community notes and professional fact-checking are intrinsically linked and their relationship is symbiotic -- professional fact-checkers perform in-depth research that amateur platform users can't, while community notes help publicise the work of fact-checkers. Therefore, we posit, the move by platforms to end their partnerships, and therefore funding, of fact-checking organisations will significantly hinder their ability to perform their basic function of fact-checking as well as their more complex investigative journalism projects that are especially necessary for community note writers. This, in turn, will significantly limit the effectiveness of community notes, specifically for high-stakes claims such as these related to broader narratives or conspiracies.

% \section*{Acknowledgements}
% This research was partially funded by a DFF Sapere Aude research leader grant under grant agreement No 0171-00034B, the Danish-Israeli Study Foundation in Memory of Josef and Regine Nachemsohn, and the Privacy Black \& White project, a UCPH Data+ Grant. This work was further supported by the Pioneer Centre for AI, DNRF grant number P1.

\section*{Limitations}
\label{sec:limitations}

The main limitations of our work concern the characteristics of the dataset we analyse. First, we restrict our analysis to notes written in English, excluding over half a million notes in other languages. This decision was made to avoid potential noise and biases arising from the authors’ unfamiliarity with public discourse in different regions and reliance on machine translation. In future work, we aim to extend our analysis to other languages.

Moreover, except for a small subset of notes, we did not have access to the original tweets they were written for. Even when the tweet text was available, many contained non-text media, were written in internet vernacular that was challenging to interpret, or lacked important context. These factors limit the accuracy and effectiveness of our models and analysis.

Finally, due to resource constraints, our manual annotation study was limited to a relatively small sample of tweets and notes. In future work, we wish to utilise crowd workers to not only annotate a larger dataset but also increase the diversity and perspective of the annotators.

\section*{Broader Impact and Ethical Considerations}
\label{sec:ethics}

% Our findings have implications for 
% firstly, the

% The Community Note dataset is a rich dataset and we hope that researchers will study it more.

Community notes have been proposed as a replacement for professional fact-checkers and a salve to some of the issues encountered by fact-checking. However, our findings support the view that neither community notes nor professional fact-checkers alone are sufficient to combat the spread of misinformation on social media. Rather, a combination of these two strategies could prove a much more effective approach to addressing the full range of false content shared online, for example, by leveraging community notes to identify new checkworthy claims for professional fact-checkers, or relying on fact-checkers' expertise to resolve disputed unpublished notes (see \citet{augenstein2025communitymoderationnewepistemology} for further recommendations in this vein). In particular, as discussed above, professional fact-checking organisations are especially vital for verifying claims related to broader narratives and conspiracy theories. Moreover, professional fact-checkers remain the only viable strategy currently available for addressing partisan issues, where community notes fall short. Finally, we highlight the potential for incorporating automated, human-in-the-loop fact-checking models to assist professional and community fact-checkers alike in reckoning with vast amounts of both human- and machine-generated content.

Given that this work analyses real-world posts, ethical concerns may arise from using this data for research purposes.
Posts from non-protected accounts and Community Notes on Twitter/X are publicly available, however, we acknowledge that they may contain sensitive personal information.
To minimise any breach of anonymity and privacy, we anonymised links to individual accounts, and we do not publicly release this information. 
We do not analyse the posts or notes by individual users, and instead examine aggregated data in the form of topics and sources cited.

Although the Community Notes dataset represents attempts to curb harmful misinformation and conspiracies, given the intense partisanship involved \citep{allen2022partisan,draws_effects_2022}, as well as the explicit content of some claims, some instances may be considered offensive.
We also acknowledge that our own perspectives and biases as authors shape the impact of our findings in certain ways.
For example, as mentioned in the previous section, we were unable to analyse non-English posts in-depth, so our conclusions are likely somewhat focused on discourse in the Anglosphere (e.g., the US, UK, Ireland, Canada, Australia, New Zealand etc.).
Furthermore, although we based our criteria for conspiracy theories on well-established sources, e.g., \href{https://apnews.com/hub/conspiracy-theories}{AP News}, \href{https://www.factcheck.org/issue/conspiracy-theories/}{FactCheck.org}, the \href{https://commission.europa.eu/strategy-and-policy/coronavirus-response/fighting-disinformation/identifying-conspiracy-theories_en}{European Commission}, and identified conspiratorial narratives from both left- and right-wing sources, our own perspectives (i.e., as scientists from Western countries) may also have impacted what we considered to be conspiracy theories.

% we didn't use crowdworkers for annotation which may have given a broader perspective.

\section*{Acknowledgements}
$\begin{array}{l}\includegraphics[width=1cm]{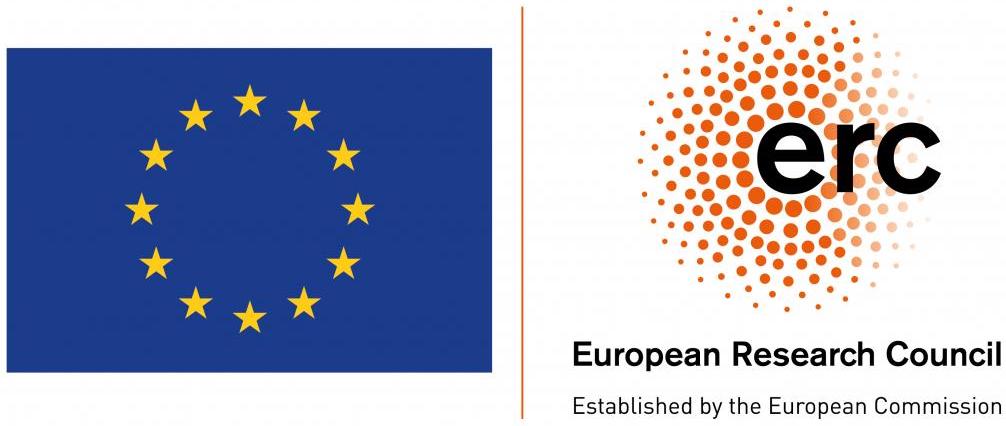} \end{array}$  This research was co-funded by the European Union (ERC, ExplainYourself, 101077481), by the European Union’s Horizon 2020 research and innovation program under grant agreement No.
101135671 (TrustLLM), and by the Pioneer Centre for AI, DNRF grant number P1.

\bibliography{anthology,custom}

\clearpage

\appendix
\section{Extended Background}
\label{app:extended_background}
\subsection{Impact of Community Notes on misinformation spread}
% Mixed evidence has emerged about the effectiveness of community notes on misinformation spread.
Posts identified by community notes as misleading have been found to attain less virality (reposts, quote tweets and replies) than non-misleading posts \cite{drolsbach_diffusion_2023,renault_collaboratively_2024}.
Community notes have also been shown to increase the probability of tweet retractions and deletions and speed up the retraction process \cite{gao_can_2024,renault_collaboratively_2024}.
However, other studies have found less positive evidence; for example, that users' followers, likes and engagement increase after their post receives a community note \cite{wirtschafter2023future}.
Curiously, one study claims that showing community notes on posts reduced the spread of misleading posts by an average of 61\% \cite{chuai_community_2024}, while a more recent analysis by the same authors found no effect of community notes on engagement with misinformation \cite{chuai_did_2024}.

%% testing effects of community notes
People shown community notes alongside misleading social media posts were more accurate in identifying misleading posts, and the notes were judged to be more trustworthy than context-free misinformation flags (e.g., "Checked by fact-checkers" or "Checked by other social media users"), regardless of (US-centric) political beliefs \cite{drolsbach2024community}.
People shown either community notes or related news article suggestions were both less likely to believe and report misleading information compared to a control group: community notes were more effective in reducing belief and sharing intention for positive rumours, while articles were more effective for negative rumours \cite{kankham_community_2024}.
% For positive rumours, people shown community notes were less likely to believe them and share them than people shown related articles, however for negative rumours, related articles were more effective in reducing self-reported belief and likelihood of sharing, although these findings were based on responses to a single health-related claim 
On the other hand, displaying community notes leads users to post more negative and angry replies to misleading posts \cite{chuai_community_2024-1}, while crowd workers are also prone to cognitive biases, such as overestimating a statement's truthfulness the more they liked its claimant, and general overconfidence in their ability to ascertain truthful statements \cite{draws_effects_2022}.

\subsection{Professional fact-checking and community note practices}
Although fact-checks and community notes share similarities in how they address misleading claims, they also differ in key elements of practice and communication \citep{kankham_community_2024}.
Fact-checking typically involves the analysis and verification of public claims, and in addition to verifying claims, in recent years many fact-checking organisations have also assumed a wider role in combating misinformation spread, e.g., long-term investigative journalism projects, media literacy programs \citep{juneja2022human}.
Professional fact-checkers signatory to the International Fact-Checking Network follow a rigorous set of principles and transparency commitments\footnote{\url{https://www.ifcncodeofprinciples.poynter.org/the-commitments}} and a structured workflow: (i) claim selection; (ii) collecting evidence; (iii) deciding on a verdict; and (iv) writing the fact-checking article \citep{graves2017anatomy,micallef2022true,warren2025explainablefactchecking}.
In contrast, any platform user can contribute to community notes under anonymity, and the rating approach relies on the `wisdom of crowds', with little oversight or transparency regarding biases of note-writers.
%rely on the 'wisdom of crowds'
Community note writers and fact-checkers tend to target similar topics (e.g., health and politics) \citep{saeed2022crowdsourced}. Fact-checkers must rely on credible sources and evidence to convince the reader, while note writers often disagree with fact-checkers on what constitutes a reliable source, particularly on political ideological grounds \cite{saeed2022crowdsourced}.
The verdicts reached in notes tend to agree with those of fact-checkers --- however, due to political polarisation \citep{yasseri2023crowdsourcing}, community notes on contentious political issues rarely reach a consensus \citep{saeed2022crowdsourced}.
Fact-checking articles, which are subject to multiple rounds of editorial scrutiny, are more formal and standardised in style than community notes, which vary considerably and can employ a range of persuasion techniques, such as appeals to emotion or other logical fallacies \citep{kankham_community_2024}.
Moreover, community notes typically serve as direct rebuttals to misleading posts, while fact-checking articles may address a more general claim than is expressed in a specific post.
Finally, fact-checking articles are a one-way exchange, while community notes represent a more horizontal and interactive dialogue between writer and recipient of the fact-check \citep{kankham_community_2024}.
Prior work has found that collaborative fact-checking, a distinct approach to community notes for its Wikipedia-style approach that allows users to edit, as well as up and downvote user-written posts \citep{Haime2022Cofacts}, can produce fact-checks with comparable speed, reliability, objectivity, clarity and persuasiveness to those written by professional fact-checks \citep{Zhao_Naaman_2023}. However, laypeople's work is expedited by existing fact-checking articles, and amateurs tend to defer to professional fact-checkers for topics requiring specific expertise, such as medical claims \citep{Zhao_Naaman_2023}.
Our work builds on current understanding of the relationship between professional fact-checking and community moderation by examining the extent to which community note writers deploy the work of fact-checkers in their notes.

\section{Additional Material}
\label{app:additional_material}

% \begin{figure}[!t]
%     \centering
%     \includegraphics[width=0.7\columnwidth, trim={0 0 0 1cm},clip]{Figures/Distribution_of_topics.png}
%     \caption{Distribution of notes' topics, with and without a fact-checking source.}
%     \label{fig:topics}
% \end{figure}

\begin{figure*}[ht]
    \centering
    \includegraphics[width=\textwidth]{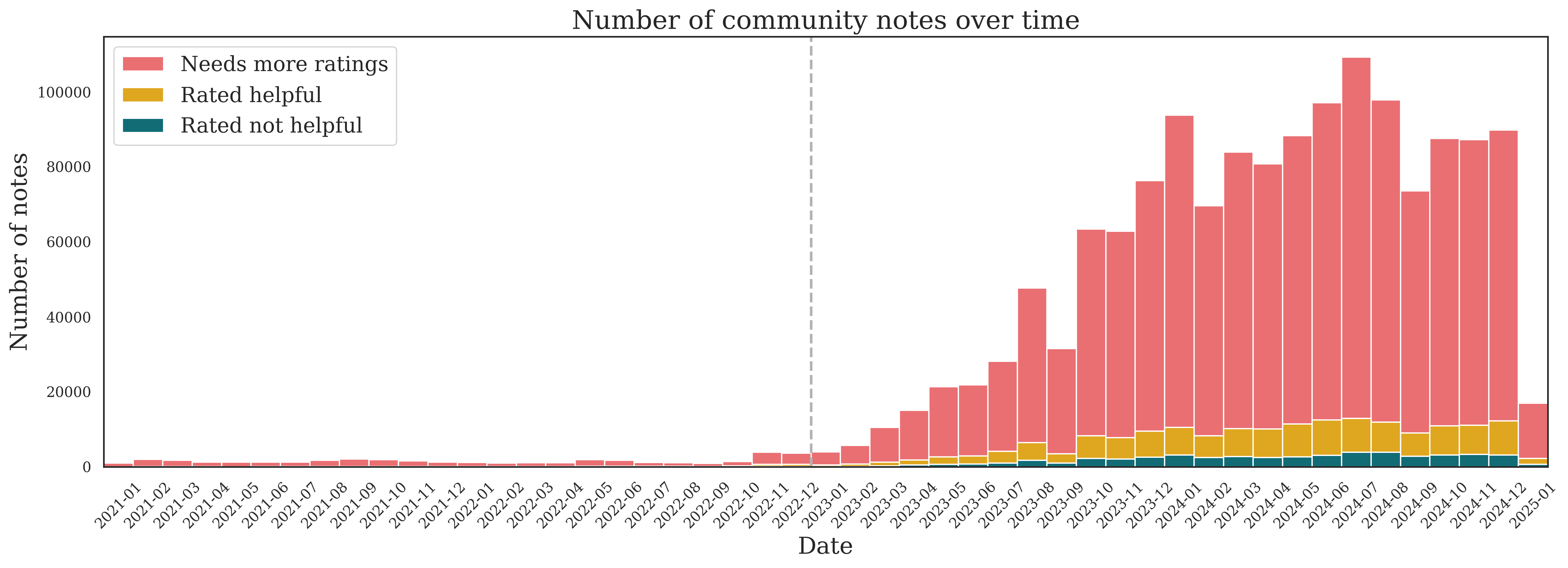}
    \caption{A histogram of the number of community notes written every month and their rating (\textit{helpful}, \textit{not helpful}, or \textit{needs more data}. The grey vertical line (December 2022) indicates the date when the community notes became visible worldwide.}
    \label{fig:notes_per_month}
\end{figure*}

% \begin{figure*}[ht]
%     \centering
%     \includegraphics[width=\textwidth]{Figures/link_types_only_helpful.png}
%     \caption{The categories of links used by Community notes' authors as a source, filtering for notes rated as ``helpful''.}
%     \label{fig:link_types_helpful}
% \end{figure*}

% \begin{figure*}[ht]
%     \centering
%     \includegraphics[width=\textwidth]{Figures/link_types_only_not_helpful.png}
%     \caption{The categories of links used by Community notes' authors as a source, filtering for notes rated as ``not helpful''.}
%     \label{fig:link_types_not_helpful}
% \end{figure*}

\begin{figure*}
    \centering
    \includegraphics[width=0.8\textwidth]{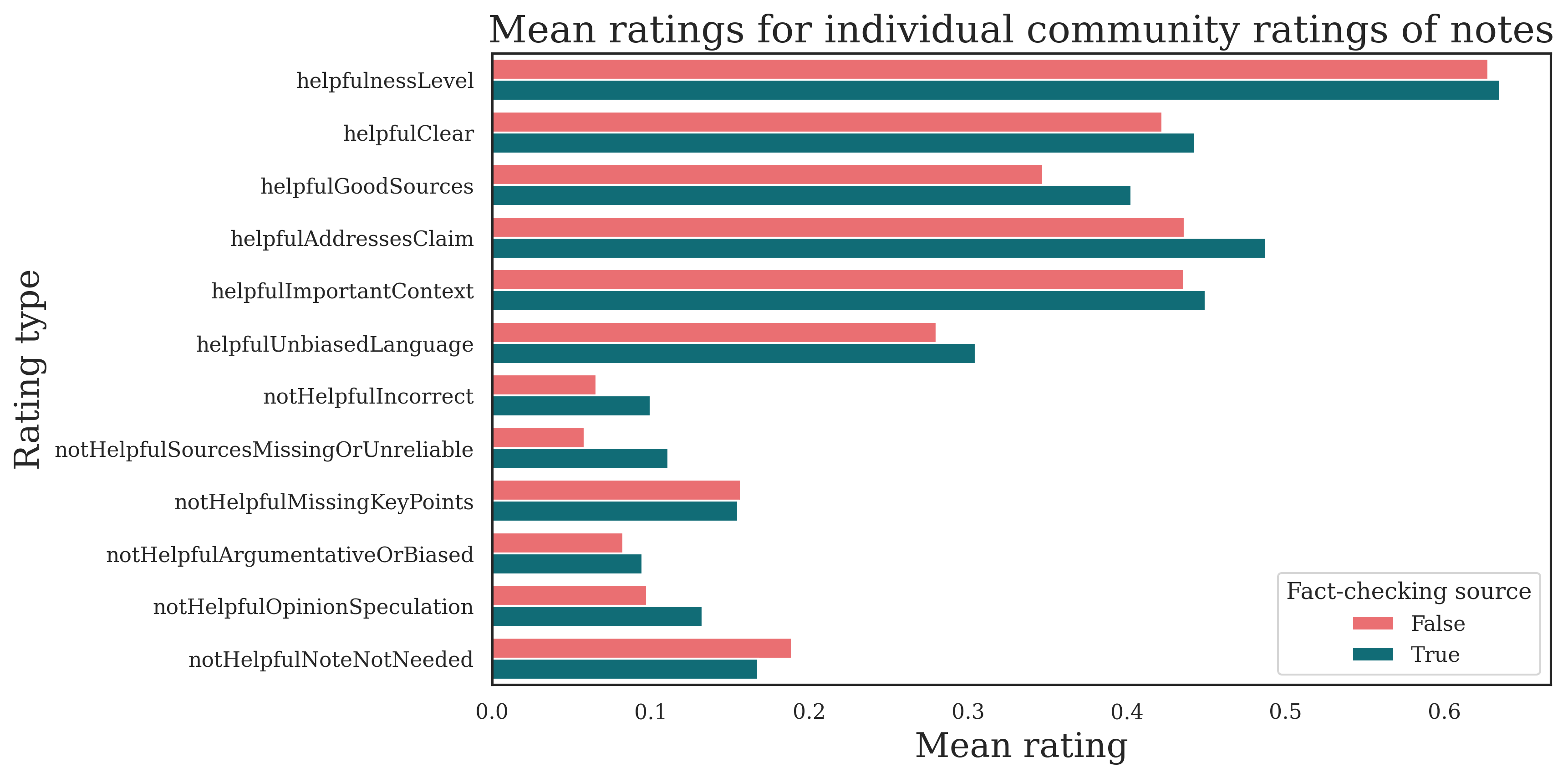}
    \caption{Community ratings of notes with and without fact-checking source.}
    \label{fig:notes_individual_ratings}
\end{figure*}

\begin{table*}
    \centering
    \resizebox{1.0\textwidth}{!}
    {%
    \fontsize{8}{8}\selectfont
    \sisetup{table-format = 3.2, group-minimum-digits=3}
    \begin{tabular}{ll|ll}
    \toprule
    \textbf{Domain} & \textbf{Category} & \textbf{Domain} & \textbf{Category} \\
    \midrule
    x.com & social media           & thehill.com & news \\
    twitter.com & social media     & amp.theguardian.com & news \\
    youtube.com & social media     & whitehouse.gov & government \\
    youtu.be & social media        & news.sky.com & news \\
    un.org & organisation          & merriam-webster.com & reference \\
    u.today & news                 & techarp.com & news \\
    t.co & social media            & cbc.ca & news \\
    snopes.com & fact checking     & politifact.com & fact checking \\
    en.m.wikipedia.org & reference & pbs.org & commercial \\
    en.wikipedia.org & reference   & telegraph.co.uk & news \\
    google.com & search engine     & businessinsider.com & news \\
    instagram.com & social media   & time.com & news \\
    britannica.com & reference     & justice.gov & government \\
    reuters.com & news             & cnbc.com & news \\
    bbc.co.uk & news               & wsj.com & news \\
    apnews.com & news              & sciencedirect.com & academic \\
    bbc.com & news                 & msn.com & news \\
    nytimes.com & news             & statista.com & reference \\
    theguardian.com & news         & business.x.com & commercial \\
    vice.com & news                & amp.cnn.com & news \\
    usatoday.com & news            & congress.gov & government \\
    factcheck.org & fact checking  & factcheck.afp.com & fact checking \\
    cnn.com & news                 & yahoo.com & search engine \\
    washingtonpost.com & news      & timesofindia.indiatimes.com & news \\
    ncbi.nlm.nih.gov & academic    & thelancet.com & academic \\
    nbcnews.com & news             & hrw.org & organisation \\
    help.twitter.com & reference   & healthfeedback.org & fact checking \\
    cdc.gov & government           & fda.gov & government \\
    npr.org & news                 & m.youtube.com & social media \\
    forbes.com & news              & law.cornell.edu & academic \\
    newsweek.com & news            & medium.com & blog post \\
    fullfact.org & fact checking   & healthfeedback.org & fact checking \\
    dailymail.co.uk & news         & who.int & organisation \\
    cbsnews.com & news             & haaretz.com & news \\
    web3antivirus.io & database    & axios.com & news \\
    timesofisrael.com & news       & mayoclinic.org & commercial \\
    help.x.com & reference         & nejm.org & academic \\
    nypost.com & news              & scienceexchange.caltech.edu & academic \\
    aljazeera.com & news           & indiatoday.in & news \\
    reddit.com & social media      & bloomberg.com & news \\
    independent.co.uk & news       & pewresearch.org & academic \\
    usgs.gov & academic            & jamanetwork.com & academic \\
    abcnews.go.com & news          & leadstories.com & news \\
    nature.com & academic          & dictionary.cambridge.org & reference \\
    gov.uk & government            & jpost.com & news \\
    web.archive.org & database     & archive.ph & database \\
    foxnews.com & news             & healthline.com & commercial \\
    tiktok.com & social media      & abc.net.au & news \\
    edition.cnn.com & news         & france24.com & news \\
    \bottomrule
    \end{tabular}
    }
    \caption{List of top 100 most common domains found in the community notes dataset, and their categorisation.}
    \label{tab:top_domains}
\end{table*}

\begin{table*}
 \resizebox{1.0\textwidth}{!}
 {%
 \fontsize{8}{8}\selectfont
 \sisetup{table-format = 3.2, group-minimum-digits=3}
    \begin{tabular}{llll}
        \toprule
        Name & URL & Language & Region/domain \\
        \midrule
        Lead stories & leadstories.com & English & Global \\
        AFP Factuel & factuel.afp.com & French & Global \\
        AAP FactCheck & aap.com.au/factcheck & English & Australia \\
        Full Fact & fullfact.org & English & Global \\
        Science Feedback & science.feedback.org & English & Science \\
        Politifact & politifact.com & English, Spanish & USA \\
        HoaxEye & hoaxeye.wordpress.com & English & Images \\
        Logically Facts & logicallyfacts.com & Multiple & Europe/India \\
        FactCheckNI & factcheckni.org & English & North Ireland \\
        DFRLab & dfrlab.org & English & Global \\
        FactReview & factreview.gr & Greek & Global \\
        Lupa & lupa.uol.com.br/jornalismo & Portuguese & Global \\
        Check your fact & checkyourfact.com & English & Global \\
        Climate feedback & climatefeedback.org & English & Climate \\
        Factcheck & factcheck.org & English & USA \\
        Health feedback & healthfeedback.org & English & Health \\
        Snopes & snopes.com & English & US \\
        aosfatos & aosfatos.org & Portuguese & Global \\
        Demagog & demagog.org.pl/fake\_news & Polish & Poland \\
        FakeReporter & fakereporter.net & Hebrew & Israel \\
        litmus factcheck & litmus-factcheck.jp & Japanese & Japan \\
        Climate Feedback & climatefeedback.org & English & Global \\
        AFP & factcheck.afp.com & English & Global \\
        USA Today & usatoday.com/story/news/factcheck & English & USA \\
        Statesman & statesman.com & English & USA \\
        Dallas News & dallasnews.com/news/politifact & English & USA \\
        Google Fact Check & toolbox.google.com/factcheck & English & Global \\
        MediaBias/FactCheck & mediabiasfactcheck.com & English & Global \\
        MedDMO & meddmo.eu & English, Greek & Greece, Cyprus, Malta \\
        Poynter & poynter.org/fact-checking & English & USA \\
        Newsmeter & newsmeter.in/fact-check & English, Tamil & India \\
        Africa Check & africacheck.org & English & Africa \\
        Fact Crescendo India & english.factcrescendo.com & English & India \\
        Factseeker & factseeker.lk & English & Sri Lanka \\
        Fact Crescendo Thailand & thailand.factcrescendo.com & Thai & Thailand \\
        Fact Crescendo Afghanistan & afghanistan.factcrescendo.com & Persian & Afghanistan \\
        Only Fact & onlyfact.in & English & India \\
        Factly & factly.in & English & India \\
        Fact Crescendo Sri Lanka & srilanka.factcrescendo.com & Sinhala & Sri Lanka \\
        Fact Crescendo Cambodia & cambodia.factcrescendo.com & Cambodian & Cambodia \\
        Becid & becid.eu & Baltic langs & Baltic \\
        Fact Hunt & facthunt.in & English & India \\
        Tec Arp & techarp.com & English & Global (based in Malaysia) \\
        10 news & 10news.com/news/fact-or-fiction & English & USA \\
        RMIT Fact Check & rmit.edu.au & English & Australia \\
        Gigafact & gigafact.org & English & USA \\
        Ayupp & ayupp.com/fact-check & English & India \\
        The Journal & thejournal.ie & English & Ireland \\
    \bottomrule
    \end{tabular}
}
\caption{List of professional fact-checking organisations and their URLs.}
\label{tab:fact_check_orgs}
\end{table*}

\begin{table*}
    \centering
    \resizebox{1.0\textwidth}{!}
    {%
    \fontsize{8}{8}\selectfont
    \sisetup{table-format = 3.2, group-minimum-digits=3}
    \begin{tabular}{p{5cm}p{7cm}rrrrr}
    \toprule
    Tweet & Note & \rotatebox[origin=r]{270}{misleadingUnverifiedClaimAsFact} & \rotatebox[origin=r]{270}{misleadingOutdatedInformation} & \rotatebox[origin=r]{270}{misleadingFactualError} & \rotatebox[origin=r]{270}{misleadingSatire} & \rotatebox[origin=r]{270}{Fact Checking source} \\ \midrule
    The NASA War Document is absolutely terrifying \url{https://t.co/...} & misrepresenting a presentation by NASA scientist Dennis Bushnell, The lecture was not detailing plans by NASA to attack the world it was a lecture for defense industry professionals, and how defense tactics might rise to meet evolving threats in the future.   \url{https://leadstories.com/hoax-alert/2021/06/fact-check-the-future-is-now-is-not-a-nasa-war-document-plan-for-world-domination-and-phasing-out-of-humans.html} & \cmark & \xmark & \xmark & \xmark & \cmark \\ \addlinespace
    BREAKING NEWS: International Criminal Investigation calls on every public citizen to recommend indictments for Bill Gates, Anthony Fauci, Pfizer, BlackRock, Tedros and Christian Drosten for pushing everyone to receive the ineffective highly dangerous lethal experimental vaccines... & Video has been fact-checked by USA Today, was found to be misleading, and promotes a conspiracy theory about COVID ... \url{https://ca.movies.yahoo.com/movies/fact-check-viral-video-promotes-204414488.html} & \cmark & \xmark & \xmark & \xmark & \cmark \\ \addlinespace
    1) California is RED.
    It is just because of the MASSIVE Election Fraud that stupid, brainwashed people believe Calif. is blue. Joe Biden won only in the SFO Bay area ... & The map shows the results of Reagan's reelection in 1984, not Biden's election in 2020.  \url{https://en.wikipedia.org/wiki/1984\_United\_States\_presidential\_election\_in\_California} & \xmark & \cmark & \xmark & \xmark & \xmark \\ \addlinespace
    Davis blows up \$100,000 fireworks in Kai Cenat setup During the Mr Beast Stream ... & The second photo is from a house fire in Atlanta in 2019. \url{https://www.11alive.com/article/news/local/woodland-brook-drive-cause-of-house-fire/85-ecb7df9b-5f65-44e9-bf9d-8c162d36c334} & \xmark & \cmark & \xmark & \xmark & \xmark \\ \addlinespace
    @cnviolations I swear community notes are the only good thing Elon added since he bought Twitter. & Community notes was first launched under former Twitter CEO Jack Dorsey in 2021 under the name of `Birdwatch'. The only thing Elon Musk did was that he renamed the feature to community notes.    \url{https://blog.twitter.com/en\_us/topics/product/2021/introducing-birdwatch-a-community-based-approach-to-misinformation}    \url{https://www.reuters.com/article/factcheck-elon-birdwatch-idUSL1N31Z2VG/} &
     \xmark & \xmark & \cmark & \xmark & \cmark \\ \addlinespace
    Thailand will become the first country to make the contract null and void, meaning that Pfizer will become responsible for all vaccine injuries ... & Thailand has no plans to void its Pfizer COVID vaccine contract, an official with the country’s National Vaccine Institute said. Thailand’s Department of Disease Control also rejected the claims as `fake news.' ...  \url{https://apnews.com/article/fact-check-covid-vaccine-pfizer-thailand-203948163859} & \xmark & \xmark & \cmark & \xmark & \cmark \\ \addlinespace
    Hilarious tweets by footballers, A thread: 1. Virgil Van Dijk [Current Liverpool Captain] \url{https://t.co/...} & Virgil Van Dijk did not tweet this, the tweet was made by a fan account in his name.    \url{https://www.pinkvilla.com/sports/fact-check-did-virgil-van-dijk-really-root-for-man-u-because-no-one-likes-liverpool-in-resurfaced-viral-tweet-1287250} & \xmark & \xmark & \xmark & \cmark & \cmark \\ \addlinespace
    Rob Reiner announces he’s on the Epstein Client List and Epstein Flight logs. What a fool! When a lawyer tells me to STFU, I STFU! \url{https://t.co/...} & This is a digitally altered photo that might be misinterpreted even if used as a joke.    The name Rob Reiner is misspelled, and the text is not on Reiner's X timeline.    \url{https://twitter.com/robreiner?t=iqu43-NszIW5oOM\_KqRSpw} & \xmark & \xmark & \xmark & \cmark & \xmark \\
    \bottomrule
    \end{tabular}
    }
    \caption{A sample of tweets, notes, and their community annotations, as well as whether the note contains a fact-checking link.}
    \label{tab:community_annotation_example}
\end{table*}

\begin{table*}

\begin{tabular}{lp{14cm}}
\toprule
ID & summary \\
\midrule
0 & This claim ruled mostly false. \url{https://www.politifact.com/factchecks/2020/may/07/facebook-posts/facebook-post-cites-doctors-widely-disputed-calcul/} \\
1 & The RedState article claims ``the shots do not stop transmission of the virus. This is false.    ''``Vaccines provide significant protection from 'getting it' – infection – and 'spreading it' – transmission – even against the delta variant.''    Source:    \url{https://www.usatoday.com/story/news/factcheck/2021/11/17/fact-check-covid-19-vaccines-protect-against-infection-transmission/6403678001/} \\
2 & There is no proof of this, the photo is real, it's not the last photo of the child.  But snoops say there is a tenuous link the parents used the same law firm to represent them as Maxwell      \url{https://www.snopes.com/fact-check/ghislaine-maxwell-jonbenet-ramsey/} \\
3 & unfounded    \url{https://www.snopes.com/fact-check/ashley-biden-diary-afraid/}   \\
4 & The mRNA vaccine does not cause cancer:  \url{https://www.factcheck.org/2024/05/still-no-evidence-covid-19-vaccination-increases-cancer-risk-despite-posts/} \\
5 & Many of the details in this popular essay are inaccurate and too numerous to list here.  The essay was fact checked by Snopes in 2005:  \url{https://www.snopes.com/fact-check/the-price-they-paid/}   \\
6 & POLITIFACT - rates False.   The report analysed a small sample of 128 temp stations out of several thousand volunteer-run stations, then extrapolated results. NOAA uses 2 programs to record daily temps. The report did not look at the 900 more sophisticated automated stations.   \url{https://www.politifact.com/factchecks/2022/aug/19/facebook-posts/fact-checking-talking-point-about-corrupted-climat/}   \\
7 & There is no verifiable evidence of campaign espionage in either the 2020 or the 2016 presidential elections.    \url{https://www.snopes.com/fact-check/obama-spying-trump-campaign/}  \url{https://www.washingtonpost.com/politics/2019/05/06/whats-evidence-spying-trumps-campaign-heres-your-guide/} \\
8 & Ladapo did get caught altering COVID vaccine study findings. Ladapo replaced the language from an earlier study draft that found no significant risk from COVID vaccines, to then state there was a high risk    \url{https://healthfeedback.org/claimreview/analysis-florida-department-health-surgeon-general-joseph-ladopo-contains-multiple-methodological-problems-covid-19-mrna-vaccines/}    \url{https://healthexec.com/topics/clinical/COVID-19/florida-surgeon-general-altered-covid-19-study-findings} \\
\bottomrule
\end{tabular}
\caption{Examples of community notes containing fact-checking sources that are rated as having \textit{notHelpfulSourcesMissingOrUnreliable}.}
\label{tab:notes_with_bad_source.}
\end{table*}

This section details additional results or material referenced from the paper's main body.

% \noindent \textbf{\Cref{fig:topics}} Distribution of notes' topics, with and without a fact-checking source.

\noindent \textbf{\Cref{fig:notes_per_month}} A histogram of the number of community notes written every month and their rating (\textit{helpful}, \textit{not helpful}, or \textit{needs more data}).

% \noindent \textbf{\Cref{fig:annotations}} Mean scores of community annotations of misleading posts.

\noindent \textbf{\Cref{fig:notes_individual_ratings}} Community ratings of notes with and without fact-checking source.

\noindent \textbf{\Cref{tab:top_domains}} List of top 100 most common domains found in the community notes dataset, and their categorisation.

\noindent \textbf{\Cref{tab:fact_check_orgs}} List of professional fact-checking organisations and their URLs.

\noindent \textbf{\Cref{tab:community_annotation_example}} A sample of tweets, notes, and their community annotations, as well as whether the note contains a fact-checking link.

\noindent \textbf{\Cref{tab:notes_with_bad_source.}} Examples of community notes containing fact-checking sources that are rated as having \textit{notHelpfulSourcesMissingOrUnreliable}.

% \noindent \textbf{\Cref{tab:community_annotation_example}} A sample of tweets, notes, and their community annotations, as well as whether the note contains a fact-checking link.

\section{Reproducibility}
\label{app:reproducibility}

\begin{figure*}[t]
\begin{lstlisting}[label=lst:prompt_link, caption=The prompt used to classify URLs into categories., numbers=none]
SYSTEM PROMPT
You are a professional IT system who has a vast knowledge of the internet and its content. Your goal is simple, but very important: Classify URLs into categories. Choose only from the provided categories!


USER PROMPT
Read the following URLs.
Your goal is to categorize each url into one of the pre-defined categories.

Chose from the following list of categories: 
Categories = 
[
    "social media",  # Social media sites like Facebook, Twitter, Youtube etc.
    "news",  # Websites of news outlets or other organisations that report current events, such as the nytimes, the guardian, etc.
    "government",  # Government agencies and organisations, as well as websites related to policies and guidelines,  such as the CDC, department of education, FDA, etc.
    "academic",  # Academic sources, journals, and magazines, such as pubmed, nature, sciencedirect, etc.
    "blog post",  # Independent blog posts about various topics, including cooking, travel, home improvement, fandom, reviews, etc.
    "fact checking",  # professional fact checking organisations
    "database",  # Public databases such as google drive, archive.com, dropbox, etc.
    "commercial",  # Webpages of commercial organisations such as BMW, Delta, Nike, etc.
    "reference",  # Public resources such as encyclopedias, dictionaries, advocacy sources, guides, DIYs, statistics, religious sources, travel information, usage guidelines, Q&As, terms of services, etc.
    "organisation",  # non-commercial and non-government organisations such as WHO, the UN, Greenpeace, LA-Lakers, etc.
    "other",  # Any other website that does not fit into one of the previous categories.
    "unknown",  # if it is impossible to determine the category of the webpage.
]

Output format example:
[
    {
        id: <ID>,
        url: <URL>,
        category: <CATEGORY>,
    }
]


URLs:
<URLS>
\end{lstlisting}
\label{prompt_link}
\end{figure*}

\begin{figure*}[t]
\begin{lstlisting}[label=lst:prompt_conspiracy, caption=The prompt used to classify tweets and notes into broader narratives and conspiracy theories., numbers=none]
SYSTEM PROMPT
You are a professional fact-checker who specializes in analyzing misinformation spread on social media. 
Your goal is to analyse a tweet and a community note written about the tweet and decide whether the tweet spread misinformation related to a known conspiracy theory or a misleading wider narrative, and if so, which one is it.



USER PROMPT
Read the following tweets and community notes written about them.\nYour goal is to analyse them and decide whether each tweet spread misinformation related to a known conspiracy theory or a similar misleading wider narrative, and if so (and only if so!), which one.
Include your reasoning. Output the results as a json file. If a tweet does not relate to a conspiracy theory or a misleading wider narrative, output "none" in the json.

- Tweets *do not* discuss a wider narrative if the misleading information is tied to a specific singular event that is not connected to major topics on the public discourse. 
They do discuss a wider narrative if the misleading information is tied to a known conspiracy theory or to major topics on the public discorse.

Chose from the following list of theories and wider narrative: 
CONSPIRACY_THEORIES = 
[
    September 11,
    October 7,
    the great replacement,
    COVID was intentionally spread,
    the COVID outbreak is fake,
    2020 election fraud,
    vaccines cause autism,
    5G towers,
    Russian invasion of Ukraine,
    flat earth,
    chemtrails,
    Q-Anon and deep state,
    Epstein files,
    Barack Obama was not born in the USA,
    Michelle Obama is a man,
    LGBT grooming,
    fluorite in the water,
    climate change,
    Holocaust denial,
    Hunter Biden and Ukraine,
    other,
]

Output format example:
[
    {
        id: <ID>,
        is_related_to_conspiracy: <True/False>,
        conspiracy: <CONSOIRACY (or None)>,
        reasoning: <REASONING>\
    }
]

    Tweets and notes: 
    <TWEETS_AND_NOTED>

\end{lstlisting}
\label{prompt_conspiracy}
\end{figure*}

\noindent \textbf{\Cref{lst:prompt_link}} The prompt used to classify URLs into categories.

\noindent \textbf{\Cref{lst:prompt_conspiracy}} The prompt used to classify tweets and notes into broader narratives and conspiracy theories.

\subsection{Manual Annotation Setup}
\label{app:manual_annotation_setup}

\begin{figure*}
    \centering
    \includegraphics[width=\textwidth]{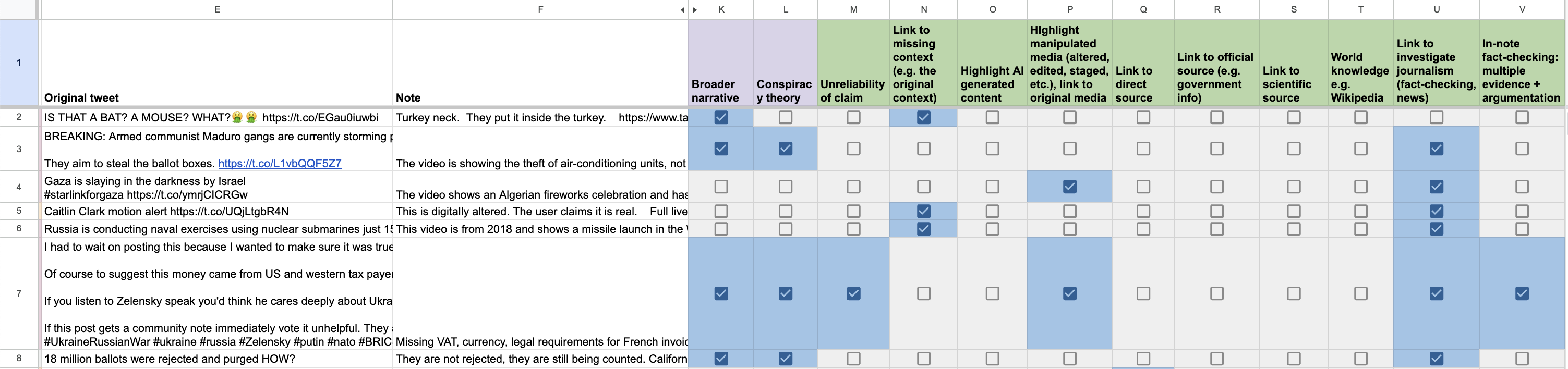}
    \caption{Our annotation setup.}
    \label{fig:annotation_setup}
\end{figure*}

We annotate 400 $(\text{tweet}, \text{note})$ pairs from $\mathcal{S}_\text{text}$ with 12 binary attributes. Each $(\text{tweet}, \text{note})$ pair was annotated in a multi-label fashion, i.e., more than one attribute can be selected at the same time. \cref{fig:annotation_setup} depict our simple annotation setup, with the 12 attributes being as follows. 

\begin{description}[topsep=0pt,itemsep=-1ex,partopsep=1ex,parsep=1ex]
    \item[Broader narrative] Whether the $(\text{tweet}, \text{note})$ pair is related to a broader narrative or a conspiracy theory.
    \item[Discredit source of claim] If the community note describes the source shared by the original post as non-credible.
    \item[Add missing context] If the community note provides some missing context to refute a claim.
    \item[Highlight AI generated] If the community note claims that the post shared AI-generated content.
    \item[Highlight edited media] If the community note claims that the post shared some media that was edited (edited with Photoshop, the clip was cut, etc.).
    \item[Link to direct source] If the community note shares a link to a source where an entity says that a claim made about them is false.
    \item[Link official source] If the community note shares a link to an official source such as a government website.
    \item[Link scientific source] If the community note shares a link to some scientific article or website.
    \item[Link world knowledge] If the community note shares a link to some reference resources such as Wikipedia.
    \item[Link fact-checking] If the community note shares a link to a professional fact-checking organisation.
    \item[In-note fact-checking] If the community note performs an in-note fact-check by cross-referencing several sources and constructing a compelling argument.
\end{description}

\end{document}